%% file: main_arxiv.tex
\documentclass[11pt]{article}

\usepackage[margin=1in]{geometry}
\usepackage[T1]{fontenc}
\usepackage{lmodern}
\usepackage{microtype}
\usepackage[section]{placeins}
\usepackage{float}
\usepackage{amsmath,amssymb,amsfonts}
\usepackage{graphicx}
\usepackage{booktabs}
\usepackage{cite}
\usepackage{bm}
\usepackage{amsthm}
\usepackage{url}
\usepackage{multirow}
\usepackage{makecell}
\usepackage{xcolor}
\usepackage[colorlinks=true,allcolors=blue!50!black]{hyperref}

\makeatletter
\providecommand\@LN@col[1]{}
\providecommand\@LN[2]{}
\makeatother

\input{macros}
\newcommand{\frameworkfigwidth}{\textwidth}
\newcommand{\misspecfigpos}{H}
\newcommand{\misspecfigwidth}{0.45\linewidth}

\begin{document}

\title{SIDUMN: Self-Supervised Image Denoising\\ Under Symmetric Multiplicative Noise}

\author{
Nyan~Htet~Naing%
\thanks{Nyan Htet Naing is currently a PhD student in the Department of Mathematics, Kyungpook National University, Korea. E-mail: nyan11@knu.ac.kr}
\and
Gyeongha~Hwang%
\thanks{G.~Hwang is the corresponding author. He is with the Department of Mathematics, Yeungnam University, Korea. E-mail: ghhwang@yu.ac.kr}%
\thanks{The work of G. Hwang has been supported by the 2025 Yeungnam University Research Grant. The authors thank Naima Naheed for helpful discussions.}
}
\date{}

\maketitle

\begin{abstract}
\input{abstract}
\end{abstract}


\input{introduction}

\input{method}

\input{experiment_ablation_conclusion}

\FloatBarrier
\bibliographystyle{unsrt}
\bibliography{references}

\end{document}

%% file: macros.tex
\newtheorem{theorem}{Theorem}

\theoremstyle{remark}

\newcommand{\xx}{\ensuremath{\mathbf{x}}}
\newcommand{\yy}{\mathbf{y}}
\newcommand{\zz}{\mathbf{z}}

\newcommand{\nn}{\mathbf{n}}
\newcommand{\mm}{\mathbf{m}}

\newcommand{\one}{\mathbf{1}}
\newcommand{\E}{\mathbb{E}}

\newcommand{\X}{\mathbb{R}^d}

\newcommand{\od}{\odot}
\newcommand{\rr}{r}
\newcommand{\norm}[1]{\left\lVert#1\right\rVert}

%% file: abstract.tex
Self-supervised denoising is challenging when the corruption is multiplicative and signal-dependent. In this letter, we propose SIDUMN, a symmetric noisier framework for self-supervised denoising under mean-one symmetric multiplicative noise. Given noisy observations and the noise law, the method samples an independent synthetic noise factor from the same law and constructs two complementary noisier images by applying the factor and its reflection about one. The denoising network is trained to predict one noisier image from the other, and at inference the reconstructions of \(K\) such antithetic pairs are averaged. We show that the population target of the training loss is the clean
conditional mean minus a multiplicative cross-term. We further prove that,
for any trained network, antithetic averaging never increases the expected
squared error of a single noisier-input reconstruction. Numerical results validate the theoretical findings and the proposed method.

%% file: introduction.tex
\section{Introduction}
Image denoising is a fundamental problem in computational imaging: given a measurement corrupted by noise, the goal is to recover the underlying clean image $\xx\in\X$. The problem becomes substantially harder when the corruption is signal-dependent, so that the noise statistics vary with the clean image itself rather than being fixed and image-independent.

A particularly important instance of signal-dependent corruption is multiplicative noise, which arises naturally in several imaging modalities. In coherent imaging systems, such as synthetic aperture radar (SAR)~\cite{oliver2004sar}, ultrasound~\cite{wagner1983speckle}, optical coherence tomography, and laser imaging, the received signal is formed by the interference of many randomly phased scattered waves. The resulting speckle acts multiplicatively on the underlying reflectivity, so that brighter regions often contain larger noise fluctuations~\cite{goodman2007speckle}. Multiplicative effects also appear in transmission imaging governed by attenuation physics, such as X-ray radiography and gated X-ray detectors used in inertial confinement fusion diagnostics, where photon statistics and detector gain scale with the measured intensity~\cite{naheed2023icf}. In such settings, a convenient model is that the observed image is the clean image scaled by a random factor with mean one that is symmetric about one, so that the observation is conditionally unbiased and positive and negative fluctuations are equally likely; we restrict attention to this symmetric model.
\begin{figure}[t]
    \centering
    \includegraphics[width=\linewidth]{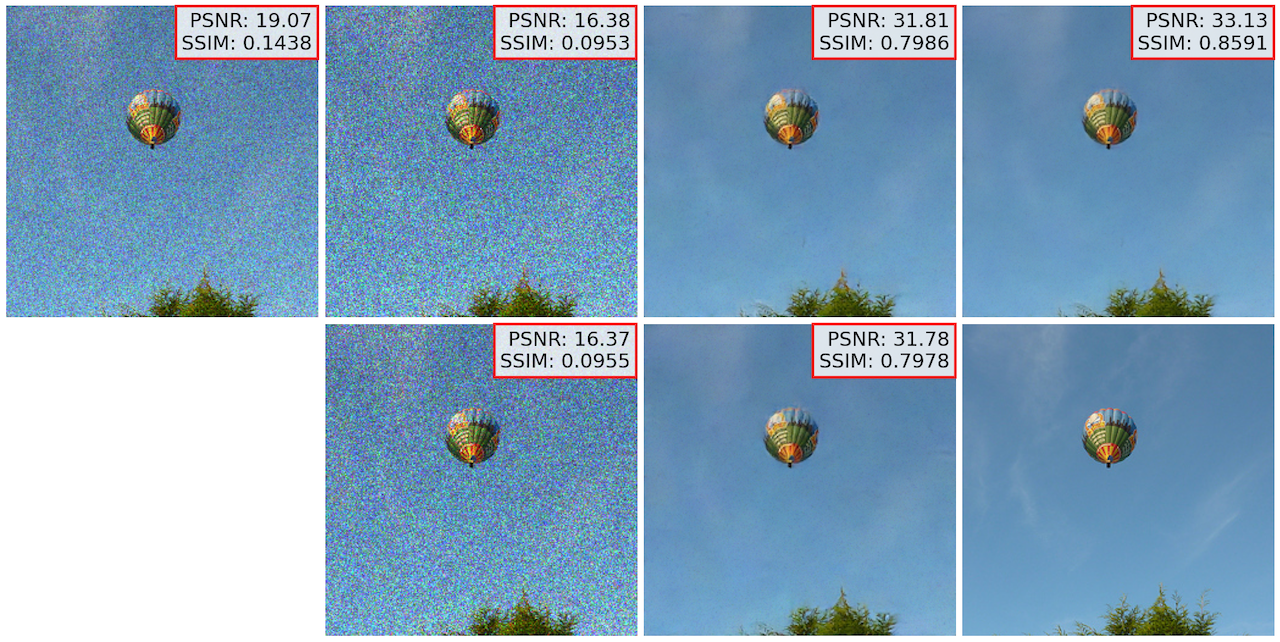}
    \caption{Illustration of SIDUMN. From left to right, the columns show the noisy image $\yy$, the complementary noisier inputs $\zz_{+}$ and $\zz_{-}$, their corresponding reconstructions $\rr_{\phi}(\zz_{+})$ and $\rr_{\phi}(\zz_{-})$, and the final antithetic average together with the clean reference image.}\label{fig:sidumn_illustration}
\end{figure}

\providecommand{\frameworkfigwidth}{0.7\textwidth}%
\begin{figure*}[t]
    \centering
    \includegraphics[width=\frameworkfigwidth]{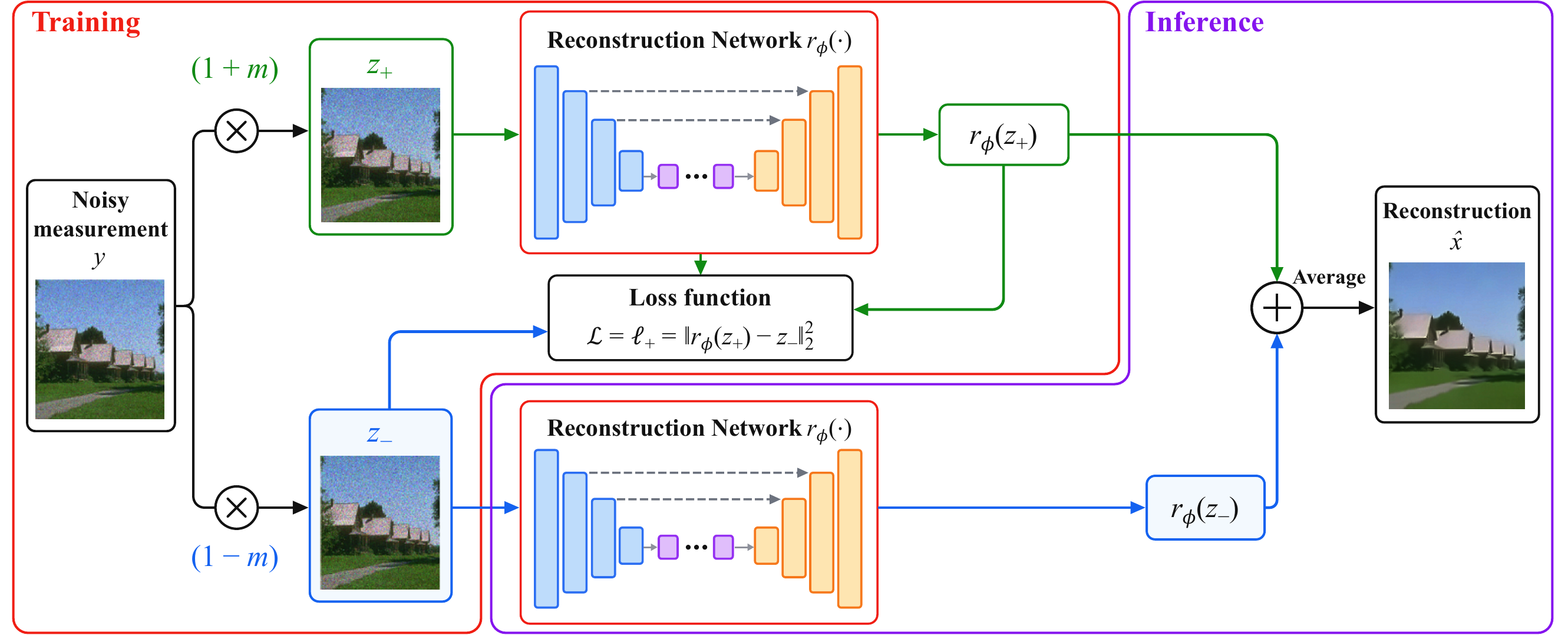}
    \caption{Details of the proposed SIDUMN framework. Given a noisy image \(\yy\), an independent synthetic multiplicative perturbation \(\mm\sim p_{\nn}\), drawn from the same mean-zero symmetric law as \(\nn\), is sampled to form two complementary noisier images \(\zz_{+}=\yy\od(\one+\mm)\) and \(\zz_{-}=\yy\od(\one-\mm)\), where \(\zz_{-}\) is the reflection of \(\zz_{+}\) about \(\yy\). The network \(\rr_{\phi}\) is trained with the loss~\eqref{eq:sidumn_loss}, which compares \(\rr_{\phi}(\zz_{+})\) with \(\zz_{-}\). At inference time, \(\rr_{\phi}\) is evaluated at both complementary inputs and the final estimate is obtained by antithetic averaging of the two complementary reconstructions.}
    \label{fig:proposed_sidumn}
\end{figure*}
Deep neural networks have achieved remarkable performance in image denoising~\cite{DnCNN, zhang2018ffdnet} when trained on paired noisy and clean images, but such pairs are often expensive or even physically impossible to obtain. Self-supervised methods instead learn directly from noisy images. Noise2Noise~\cite{Noise2Noise} demonstrated that denoising networks can be trained without clean targets by leveraging two independently corrupted observations of the same signal. To remove the need for paired noisy observations, blind-spot methods such as Noise2Self~\cite{Noise2Self}, Noise2Void~\cite{Noise2Void}, and their extensions~\cite{laine2019high, wang2022blind2unblind} exploit the 
spatial independence of the noise through masked-pixel prediction or blind-spot network architectures. Alternatively, Noisier2Noise~\cite{NN2Noise} trained a network to predict a noisy image from a noisier one formed by adding synthetic noise from a known noise model, and estimated the clean image through a correction of the network output. Building upon this, Recorrupted-to-Recorrupted~\cite{R2R} introduced a formulation that eliminates the need for such correction under Gaussian noise. More recently, Generalized Recorrupted-to-Recorrupted~\cite{GR2R} further extended the recorruption framework to accommodate a broader class of noise distributions, including additive and exponential-family noise models. Related self-supervised formulations that operate in the measurement domain for inverse problems~\cite{N2Inverse,NN2I,SSDU,Sure} have also been proposed. However, most recorruption-based methods are designed for additive noise or specific exponential-family noise models, rather than general symmetric multiplicative noise. Conversely, deep methods designed specifically for multiplicative noise have been developed mainly for SAR despeckling, with a supervised CNN~\cite{chierchia2017sar}, a Noise2Noise-style method~\cite{dalsasso2021sar2sar}, and a self-supervised blind-spot network~\cite{molini2021speckle2void}.

In this letter, we propose SIDUMN, a self-supervised denoising method for mean-one symmetric multiplicative noise. From each noisy image, SIDUMN forms two complementary noisier images with synthetic factors $\one+\mm$ and $\one-\mm$, trains a network to predict one from the other, and averages the network outputs on both at inference (Fig.~\ref{fig:sidumn_illustration}).

Our contributions are:
\begin{itemize}
\item We propose SIDUMN, which combines a symmetric synthetic perturbation strategy for noisier training with antithetic averaging inference that averages the network outputs over $K$ antithetic pairs.
\item We characterize the population target of the training loss, which contains a multiplicative cross-term. We further prove that, for any trained network, antithetic averaging never increases the expected squared error of a single noisier-input reconstruction.
\item We empirically validate SIDUMN against self-supervised baselines and analyze the effect of $K$ and of a mismatched noise level.
\end{itemize}

%% file: method.tex
\section{Methodology} \label{sec:methodology}

\begin{figure*}[t]
    \centering
    \includegraphics[width=\textwidth]{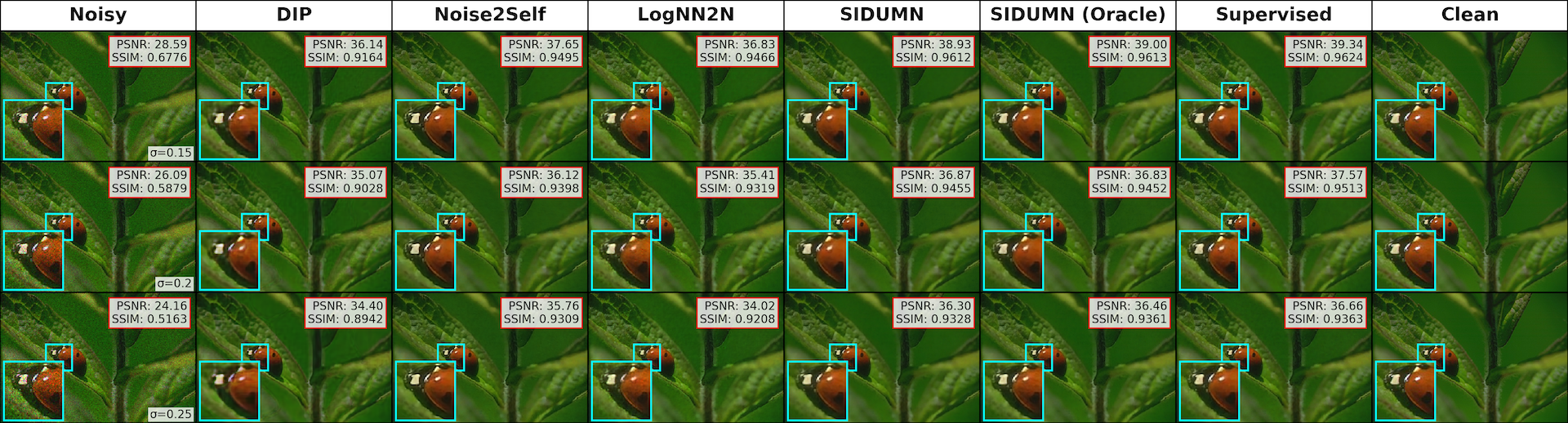}
    \includegraphics[width=\textwidth]{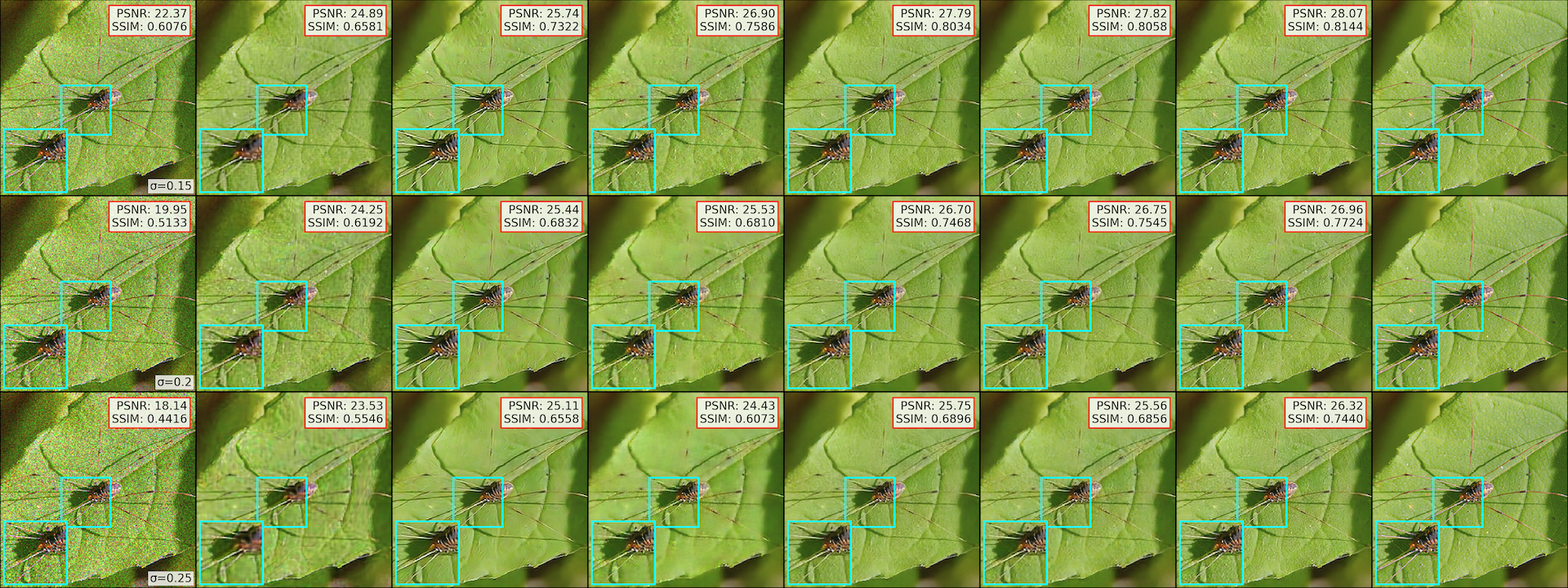}
    \caption{Representative denoising results on BSD300 (top) and ImageNet (bottom). In each block, the rows correspond to \(\sigma=0.15\), \(0.2\), and \(0.25\). Red boxes give the PSNR/SSIM of each reconstruction, and cyan boxes show zoomed-in regions.}
    \label{fig:denoising_comparison}
\end{figure*}

Our goal is to train a denoising network $\rr_{\phi}:\X\to\X$ from noisy observations $\yy$, without paired clean targets $\xx$. The overview of the method is shown in Fig.~\ref{fig:proposed_sidumn}. We consider the multiplicative observation model
\[
    \yy=\xx\od(\one+\nn), \qquad \nn\sim p_{\nn},
\]
where $\od$ denotes the element-wise product, $\nn$ is independent of $\xx$, and $p_{\nn}$ is symmetric about zero. Thus, $\E[\nn]=0$, the factor $\one+\nn$ has mean one, and $\E[\yy\mid\xx]=\xx$. We draw $\mm\sim p_{\nn}$ independently of $(\xx,\nn)$ and form
\[
    \zz_{+}=\yy\od(\one+\mm), \qquad
    \zz_{-}=\yy\od(\one-\mm).
\]
By construction, the joint distribution of $(\xx,\nn,\mm)$ is invariant
under swapping $\nn$ and $\mm$ and under flipping the sign of $\mm$:
\begin{equation}
    (\xx,\mm,\nn)\overset{\mathrm{law}}{=}(\xx,\nn,\mm)\overset{\mathrm{law}}{=}(\xx,\nn,-\mm).
    \label{eq:invariances}
\end{equation}
The first equality holds because $\nn$ and $\mm$ are i.i.d.\ and independent of $\xx$; the second holds because $\mm$ is symmetric and independent of $(\xx,\nn)$. We train $\rr_{\phi}$ using the loss
\begin{equation}
    \mathcal L_{\mathrm{SIDUMN}}(\phi)
    =
    \E_{\yy,\mm}
    \left[
        \norm{\rr_{\phi}(\zz_{+})-\zz_{-}}_2^2
    \right].
    \label{eq:sidumn_loss}
\end{equation}
The loss requires no clean image, only the noisy observation $\yy$ and
knowledge of the noise law $p_{\nn}$. The next result shows that its population minimizer is the clean conditional expectation $\E[\xx\mid\zz_{+}]$ minus a multiplicative cross-term.

\begin{theorem}\label{thm:stochastic_interpretation}
Assume all random vectors are square-integrable. Over all measurable functions, a minimizer of~\eqref{eq:sidumn_loss} satisfies $\rr^{\ast}(\zz_{+})=\E[\zz_{-}\mid\zz_{+}]$ almost surely, and
\begin{equation}
    \E[\zz_{-}\mid \zz_{+}]
    =
    \E[\xx\mid \zz_{+}]
    -
    \E[\xx\od\nn\od\mm\mid \zz_{+}].
    \label{eq:forward_bias}
\end{equation}
\end{theorem}

\begin{proof}
The first claim is the $L^2$-optimality of the conditional expectation. Expanding the target gives
\[
    \zz_{-}
    =\xx+\xx\od\nn-\xx\od\mm-\xx\od\nn\od\mm.
\]
By the first equality in~\eqref{eq:invariances}, which fixes $\zz_{+}$, $\E[\xx\od\nn\mid\zz_{+}]=\E[\xx\od\mm\mid\zz_{+}]$, so the two middle terms cancel in conditional expectation.
\end{proof}

A single
prediction $r_\phi(\mathbf{z}_+)$ depends on the draw of $\mathbf{m}$.
To reduce this variability, we propose \emph{antithetic averaging
inference}. Since $\mathbf{z}_+$ and $\mathbf{z}_-$ are exchangeable by the
second equality in \eqref{eq:invariances}, we evaluate $r_\phi$ at both complementary inputs
and average the two reconstructions,
\[
  \widehat{\mathbf{x}}
  = \tfrac12\bigl[
      r_\phi(\mathbf{z}_+) + r_\phi(\mathbf{z}_-)
    \bigr].
\]
The following theorem establishes that this averaging never increases
the expected squared error relative to a single reconstruction.

\begin{theorem}\label{thm:antithetic}
Let $\rr_{\phi}$ be any network with finite output second moments, and let $u=\rr_{\phi}(\zz_{+})$, $v=\rr_{\phi}(\zz_{-})$, and $R(\phi):=\E\norm{u-\xx}_2^2$. Then $\E\norm{v-\xx}_2^2=R(\phi)$ and
\[
    \E\norm{\tfrac12(u+v)-\xx}_2^2
    =R(\phi)-\tfrac14\E\norm{u-v}_2^2
    \le R(\phi).
\]
\end{theorem}

\begin{proof}
By the second equality in~\eqref{eq:invariances}, $(\xx,u)\overset{\mathrm{law}}{=}(\xx,v)$, which gives the first claim. Taking expectations in the identity $\norm{\tfrac12(u+v)-\xx}_2^2=\tfrac12\norm{u-\xx}_2^2+\tfrac12\norm{v-\xx}_2^2-\tfrac14\norm{u-v}_2^2$ yields the second.
\end{proof}

The single-pair estimate $\widehat{\xx}$ still depends on the draw of $\mm$. To reduce this residual Monte Carlo variance, we draw $K$ i.i.d.\ synthetic perturbations $\mm^{(1)},\ldots,\mm^{(K)}\sim p_{\nn}$, independent of $\yy$, form $\zz_{\pm}^{(k)}=\yy\od(\one\pm\mm^{(k)})$, and extend the average to $2K$ reconstructions:
\begin{equation}
    \widehat{\xx}_{\mathrm{SIDUMN}}
    =
    \frac{1}{2K}\sum_{k=1}^{K}
    \left[
        \rr_{\phi}(\zz_{+}^{(k)})+\rr_{\phi}(\zz_{-}^{(k)})
    \right].
    \label{eq:sidumn_inference}
\end{equation}
This is a Monte Carlo estimate of $\E_{\mm}[\rr_{\phi}(\yy\od(\one+\mm))]$ that pairs each $\mm^{(k)}$ with its antithetic counterpart $-\mm^{(k)}$~\cite{hammersley1956antithetic}. It requires no additional training, and Section~\ref{sec:ablation} shows that PSNR and SSIM improve monotonically with $K$.

The proposed symmetric noisier training and antithetic averaging inference
can be extended to inverse problems $\yy=(A\xx)\od(\one+\nn)$ with a known
forward operator $A$. With $\zz_{\pm}=\yy\od(\one\pm\mm)$ formed as before,
we train a reconstruction network $\rr_{\phi}$ by minimizing
$\E\norm{A\rr_{\phi}(\zz_{+})-\zz_{-}}_2^2.$ Then~\eqref{eq:forward_bias}
holds in the measurement domain with $\xx$ replaced by $A\xx$ and Theorem~\ref{thm:antithetic} holds
unchanged. Inference uses~\eqref{eq:sidumn_inference} exactly as in denoising.

%% file: experiment_ablation_conclusion.tex
\section{Numerical Experiments}

We synthesize noisy images from two clean-image datasets: BSD300~\cite{bsd} (\(300\) images, split into \(200\) training, \(50\) validation, and \(50\) test images) and \(6{,}000\) ImageNet~\cite{ILSVRC15} validation images (split into \(5{,}000\), \(500\), and \(500\) images, respectively). We generate \(\yy=\xx\od(\one+\nn)\) with \(\nn\sim\mathcal N(\mathbf 0,\sigma_n^2 I)\) without clipping, and draw \(\mm\) from the same law, so the assumptions of Section~\ref{sec:methodology} hold exactly. 

We compare SIDUMN with three self-supervised denoisers, DIP~\cite{DIP}, Noise2Self~\cite{Noise2Self}, and log-domain Noisier2Noise (LogNN2N)~\cite{NN2Noise, NN2I}, and with a supervised baseline trained with the squared \(\ell_2\) loss on noisy--clean image pairs. All methods except DIP use the U-Net~\cite{UNet} architecture. DIP\footnote{\url{https://github.com/DmitryUlyanov/deep-image-prior}} and Noise2Self\footnote{\url{https://github.com/czbiohub-sf/noise2self}} follow their public implementations; DIP is optimized independently on each test image. Like SIDUMN, DIP and Noise2Self denoise \(\yy\) directly in the intensity domain, whereas LogNN2N works in the log domain; together, they cover the blind-spot and recorruption designs underlying many self-supervised speckle denoisers. LogNN2N minimizes
\[
    \mathcal L_{\log}(\phi)
    :=
    \E
    \left[
        \norm{\rr_\phi(\tilde{\zz})-(2\tilde{\yy}-\tilde{\zz})}_2^2
    \right],
\]
where \(\tilde{\yy}=\log\max(\yy,\epsilon)\), \(\tilde{\zz}=\log\max(\zz_{+},\epsilon)\), with the maximum taken elementwise and \(\epsilon=10^{-6}\) to handle nonpositive values, and \(\widehat{\xx}=\exp(\rr_\phi(\tilde{\yy}))\). All models except DIP are trained for \(5{,}000\) epochs on BSD300 and \(500\) on ImageNet with Adam~\cite{ADAM} (learning rate \(10^{-5}\), batch size \(4\)). For the other baselines and SIDUMN (Oracle), we report the checkpoint with the highest validation PSNR, and for DIP the iterate with the highest PSNR on each test image; both use clean images. SIDUMN uses no clean images and is reported at the last training epoch. Both SIDUMN variants use \(K=32\). PSNR and SSIM~\cite{ssim_paper} are computed against the clean images.

\begin{table}[t]
\centering
\caption{Denoising results on BSD300 and ImageNet: PSNR (dB) / SSIM; best and second-best self-supervised results, excluding SIDUMN (Oracle), are bold and italic. SIDUMN (Oracle) uses the checkpoint with the highest validation PSNR against clean images.}
\label{tab:main_results}
\footnotesize
\setlength{\tabcolsep}{3pt}
\renewcommand{\arraystretch}{1.02}
\begin{tabular}{@{}lccc@{}}
\toprule
\textbf{BSD300}
& \(\boldsymbol{\sigma=0.15}\)
& \(\boldsymbol{\sigma=0.2}\)
& \(\boldsymbol{\sigma=0.25}\) \\
\midrule
Noisy input
& 23.40\,/\,0.6269
& 21.00\,/\,0.5440
& 19.17\,/\,0.4763 \\
DIP~\cite{DIP}
& 28.49\,/\,0.8179
& 27.68\,/\,0.7924
& 26.96\,/\,0.7683 \\
Noise2Self~\cite{Noise2Self}
& 30.00\,/\,\emph{0.8977}
& \emph{29.69}\,/\,\emph{0.8783}
& \emph{29.06}\,/\,\textbf{0.8597} \\
LogNN2N~\cite{NN2Noise, NN2I}
& \emph{30.92}\,/\,0.8860
& 29.16\,/\,0.8469
& 27.83\,/\,0.8153 \\
SIDUMN
& \textbf{32.23}\,/\,\textbf{0.9164}
& \textbf{30.71}\,/\,\textbf{0.8868}
& \textbf{29.23}\,/\,\emph{0.8412} \\
\midrule
SIDUMN (Oracle)
& 32.29\,/\,0.9187
& 30.78\,/\,0.8899
& 29.43\,/\,0.8520 \\
Supervised
& 32.54\,/\,0.9227
& 31.07\,/\,0.8994
& 29.97\,/\,0.8744 \\
\bottomrule
\end{tabular}
\par\vspace{3pt}
\begin{tabular}{@{}lccc@{}}
\toprule
\textbf{ImageNet}
& \(\boldsymbol{\sigma=0.15}\)
& \(\boldsymbol{\sigma=0.2}\)
& \(\boldsymbol{\sigma=0.25}\) \\
\midrule
Noisy input
& 23.15\,/\,0.6422
& 20.79\,/\,0.5669
& 18.98\,/\,0.5053 \\
DIP~\cite{DIP}
& 27.83\,/\,0.7951
& 26.99\,/\,0.7686
& 26.26\,/\,0.7439 \\
Noise2Self~\cite{Noise2Self}
& 28.88\,/\,0.8792
& 28.87\,/\,\emph{0.8594}
& \emph{28.38}\,/\,\textbf{0.8406} \\
LogNN2N~\cite{NN2Noise, NN2I}
& \emph{30.35}\,/\,\emph{0.8859}
& \emph{29.12}\,/\,0.8580
& 28.02\,/\,\emph{0.8318} \\
SIDUMN
& \textbf{31.55}\,/\,\textbf{0.9021}
& \textbf{29.99}\,/\,\textbf{0.8645}
& \textbf{28.70}\,/\,0.8218 \\
\midrule
SIDUMN (Oracle)
& 31.57\,/\,0.9034
& 30.00\,/\,0.8672
& 28.58\,/\,0.8343 \\
Supervised
& 31.86\,/\,0.9122
& 30.50\,/\,0.8879
& 29.62\,/\,0.8700 \\
\bottomrule
\end{tabular}
\end{table}

Table~\ref{tab:main_results} shows that SIDUMN achieves the best PSNR among the self-supervised methods at every noise level on both datasets, ahead of the strongest baseline by \(0.2\)--\(1.3\) dB on BSD300 and \(0.3\)--\(1.2\) dB on ImageNet. It also gives the best SSIM except at \(\sigma=0.25\), where Noise2Self (and, on ImageNet, LogNN2N) is higher. Although it uses no clean images for training or checkpoint selection, SIDUMN is within \(0.2\) dB of the oracle-selected checkpoint and \(0.3\)--\(0.9\) dB below the supervised model, so the lack of clean validation data costs little.

Fig.~\ref{fig:denoising_comparison} shows one reconstruction example from each dataset. DIP and LogNN2N over-smooth fine structures, whereas SIDUMN preserves more of them while removing the noise and gives the highest PSNR and SSIM among the self-supervised methods in every row.

\section{Ablation Study} \label{sec:ablation}
We compare inference rules and then examine a limitation: sensitivity to a mismatch between the true and synthetic noise levels.

\textbf{Inference rule.}
Table~\ref{tab:ablation} evaluates the same trained SIDUMN network with three inference rules: direct inference \(\rr_\phi(\yy)\), a single noisier input \(\rr_\phi(\zz_+)\), and the antithetic average~\eqref{eq:sidumn_inference} with \(K\in\{1,2,4,8,16,32\}\) pairs. A single noisier input gives the lowest PSNR, as \(\zz_+\) is more corrupted than \(\yy\), except on BSD300 at \(\sigma=0.25\), where it matches direct inference. Averaging one antithetic pair (\(K=1\)) improves PSNR over \(\rr_\phi(\zz_+)\), consistent with Theorem~\ref{thm:antithetic}, and also over direct inference \(\rr_\phi(\yy)\) on both datasets: the errors of the two complementary reconstructions partly cancel. More pairs give a further, monotone gain that grows with the noise level; from \(K=1\) to \(K=32\), PSNR rises by \(0.06\)--\(0.23\) dB on BSD300 and \(0.10\)--\(0.35\) dB on ImageNet, and SSIM by up to \(0.022\).
\begin{table}[t]
\centering
\caption{Ablation on the inference rule: test-set PSNR (dB) / SSIM; best in bold.}
\label{tab:ablation}
\footnotesize
\setlength{\tabcolsep}{3pt}
\renewcommand{\arraystretch}{0.92}
\begin{tabular}{@{}lccc@{}}
\toprule
\textbf{BSD300}
& \(\boldsymbol{\sigma=0.15}\)
& \(\boldsymbol{\sigma=0.2}\)
& \(\boldsymbol{\sigma=0.25}\) \\
\midrule
\(\rr_\phi(\yy)\)   & 31.46\,/\,0.8921 & 29.60\,/\,0.8437 & 27.74\,/\,0.7652 \\
\(\rr_\phi(\zz_+)\) & 30.76\,/\,0.8929 & 29.28\,/\,0.8534 & 27.74\,/\,0.7858 \\
Avg. \((K=1)\)      & 32.17\,/\,0.9148 & 30.61\,/\,0.8825 & 29.00\,/\,0.8275 \\
Avg. \((K=2)\)      & 32.20\,/\,0.9156 & 30.66\,/\,0.8847 & 29.12\,/\,0.8343 \\
Avg. \((K=4)\)      & 32.21\,/\,0.9160 & 30.69\,/\,0.8858 & 29.17\,/\,0.8379 \\
Avg. \((K=8)\)      & 32.22\,/\,0.9163 & 30.70\,/\,0.8864 & 29.20\,/\,0.8398 \\
Avg. \((K=16)\)     & 32.22\,/\,\textbf{0.9164} & \textbf{30.71}\,/\,0.8866 & 29.22\,/\,0.8407 \\
Avg. \((K=32)\)     & \textbf{32.23}\,/\,\textbf{0.9164} & \textbf{30.71}\,/\,\textbf{0.8868} & \textbf{29.23}\,/\,\textbf{0.8412} \\
\bottomrule
\end{tabular}
\par\vspace{3pt}
\begin{tabular}{@{}lccc@{}}
\toprule
\textbf{ImageNet}
& \(\boldsymbol{\sigma=0.15}\)
& \(\boldsymbol{\sigma=0.2}\)
& \(\boldsymbol{\sigma=0.25}\) \\
\midrule
\(\rr_\phi(\yy)\)   & 30.56\,/\,0.8720 & 28.75\,/\,0.8156 & 27.47\,/\,0.7586 \\
\(\rr_\phi(\zz_+)\) & 30.15\,/\,0.8768 & 28.48\,/\,0.8183 & 27.05\,/\,0.7540 \\
Avg. \((K=1)\)      & 31.45\,/\,0.8988 & 29.78\,/\,0.8532 & 28.35\,/\,0.7997 \\
Avg. \((K=2)\)      & 31.50\,/\,0.9005 & 29.89\,/\,0.8589 & 28.52\,/\,0.8106 \\
Avg. \((K=4)\)      & 31.52\,/\,0.9014 & 29.94\,/\,0.8619 & 28.62\,/\,0.8164 \\
Avg. \((K=8)\)      & 31.54\,/\,0.9018 & 29.97\,/\,0.8634 & 28.66\,/\,0.8195 \\
Avg. \((K=16)\)     & 31.54\,/\,0.9020 & 29.98\,/\,0.8642 & 28.69\,/\,0.8210 \\
Avg. \((K=32)\)     & \textbf{31.55}\,/\,\textbf{0.9021} & \textbf{29.99}\,/\,\textbf{0.8645} & \textbf{28.70}\,/\,\textbf{0.8218} \\
\bottomrule
\end{tabular}
\end{table}
\providecommand{\misspecfigpos}{t}%
\providecommand{\misspecfigwidth}{0.49\linewidth}
\begingroup\edef\x{\endgroup\noexpand\begin{figure}[\misspecfigpos]}\x
    \centering
    \includegraphics[width=\misspecfigwidth]{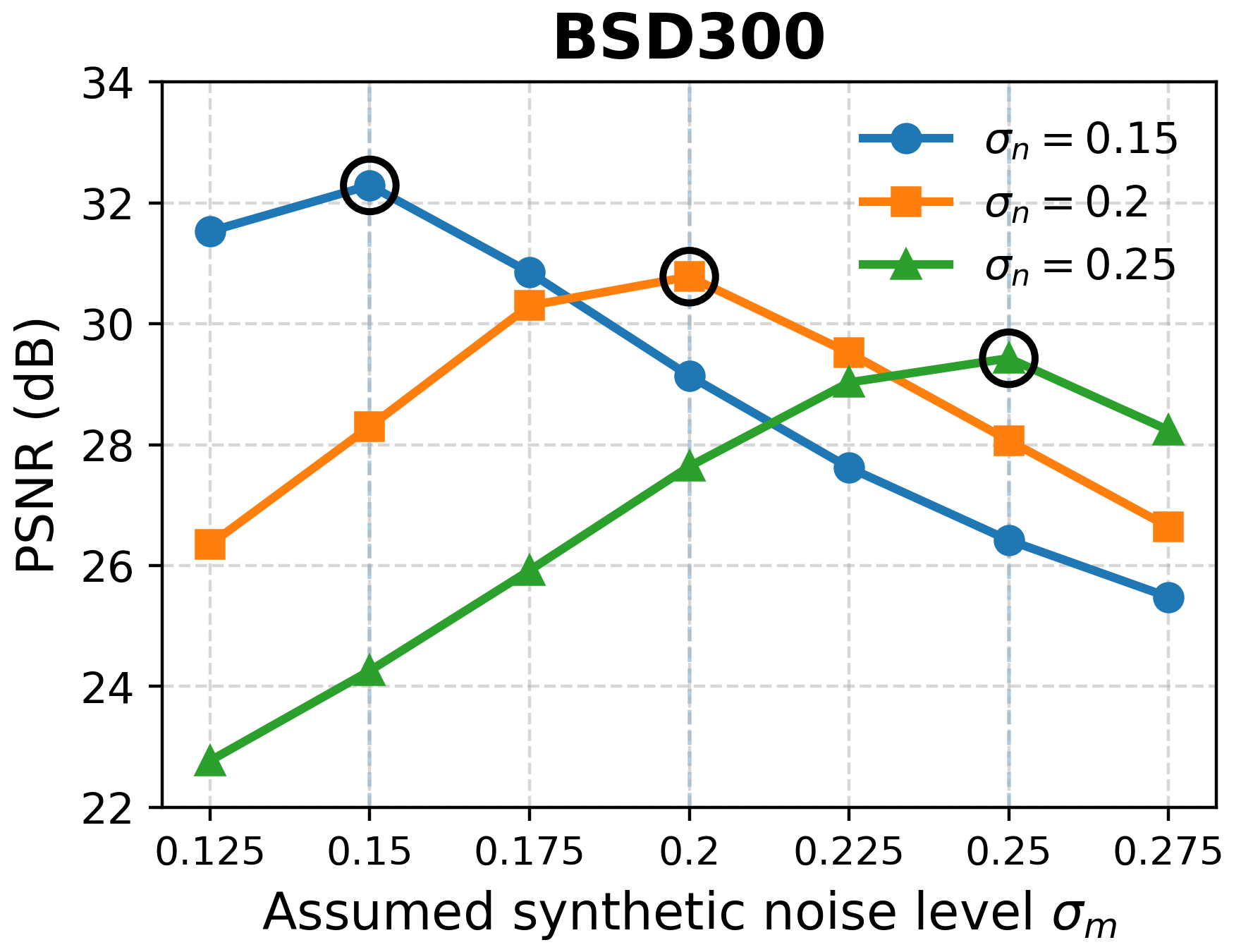}\hfill
    \includegraphics[width=\misspecfigwidth]{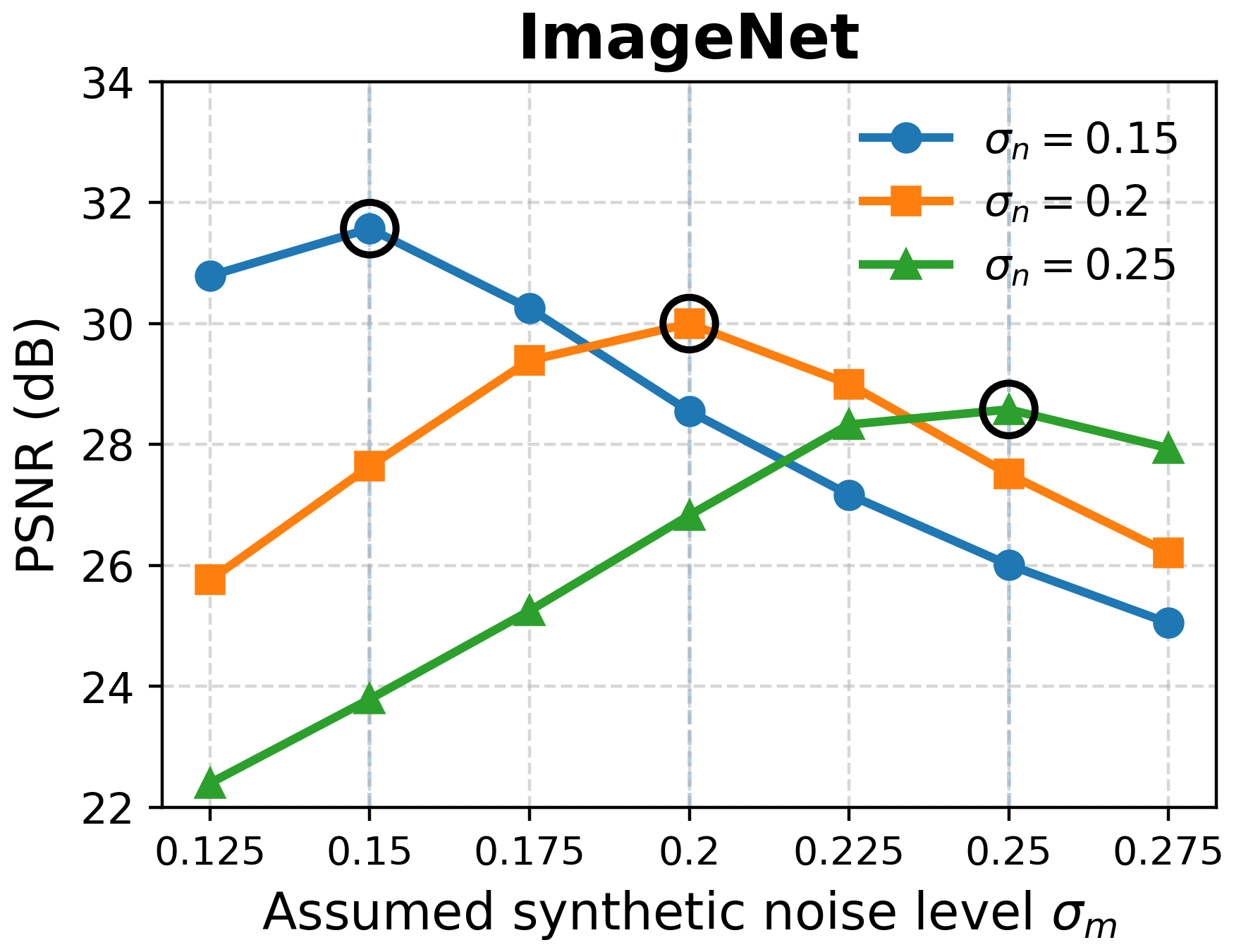}
    \caption{Noise-level misspecification (oracle-selected checkpoints): PSNR versus the assumed level \(\sigma_m\) for each true level \(\sigma_n\).}
    \label{fig:misspecification}
\end{figure}

\textbf{Noise-level misspecification.}
Since the true level \(\sigma_n\) may be unknown in practice, we train and test SIDUMN with synthetic levels \(\sigma_m\in\{0.125,0.15,\ldots,0.275\}\) for each \(\sigma_n\in\{0.15,0.2,0.25\}\). On both datasets, PSNR peaks at \(\sigma_m=\sigma_n\) among the tested levels and decreases as the mismatch grows (Fig.~\ref{fig:misspecification}). A one-step mismatch (\(\pm0.025\)) costs \(0.3\)--\(1.4\) dB, and overestimating \(\sigma_m\) by one step costs more than underestimating it.

\section{Conclusion}

We proposed SIDUMN, a self-supervised denoiser for mean-one symmetric multiplicative noise that combines symmetric noisier training with antithetic averaging inference. We characterized the population target of the loss and proved that antithetic averaging never increases the expected squared error of a single noisier-input reconstruction. On BSD300 and ImageNet, SIDUMN gives the best PSNR among self-supervised methods, close to the oracle and supervised references. Future work includes estimating the noise level jointly with training and validating SIDUMN experimentally on inverse problems.